\documentclass[11pt, a4paper, logo, twocolumn, copyright, nonumbering]{googledeepmind}

\pdftrailerid{redacted}

\makeatletter
\renewcommand\bibentry[1]{\nocite{#1}{\frenchspacing\@nameuse{BR@r@#1\@extra@b@citeb}}}
\makeatother

\usepackage{kantlipsum, lipsum}
\usepackage{dsfont}
\usepackage{gdm-colors}
\usepackage[many]{tcolorbox}
\usepackage{xspace}
\usepackage{todonotes}

\usepackage{stackrel}

\NewTColorBox{NewBox}{ s O{tbp} }{%
  floatplacement={#2},
  IfBooleanTF={#1}{float*,width=\textwidth}{float},
  colback=red!5!white,colframe=red!70!black,%
}
 
\newcommand{\Unun}{{\color{red}Un}\textit{Unlearning}\xspace}

\newcommand{\ununing}{{\color{red}un}\textit{unlearning}\xspace}

\usepackage[authoryear, sort&compress, round]{natbib}

\graphicspath{{figures/}}

\title{\Unun: Unlearning is not sufficient for content regulation in advanced generative AI}

\correspondingauthor{iliashumailov@google.com}

\renewcommand{\today}{}

\author[1]{Ilia Shumailov}
\author[1]{Jamie Hayes}
\author[1]{Eleni Triantafillou}
\author[1]{Guillermo Ortiz-Jimenez}
\author[1]{Nicolas Papernot}
\author[1]{Matthew Jagielski}
\author[1]{Itay Yona}
\author[1]{Heidi Howard}
\author[1]{Eugene Bagdasaryan}

\affil[1]{Google DeepMind}

\begin{abstract}
Exact unlearning was first introduced as a privacy mechanism that allowed a user to retract their data from machine learning models on request. Shortly after, inexact schemes were proposed to mitigate the impractical costs associated with exact unlearning. More recently unlearning is often discussed as an approach for removal of impermissible knowledge i.e. knowledge that the model should not possess such as unlicensed copyrighted, inaccurate, or malicious information. The promise is that if the model does not have a certain malicious capability, then it cannot be used for the associated malicious purpose. In this paper we revisit the paradigm in which unlearning is used for in Large Language Models (LLMs) and highlight an underlying inconsistency arising from in-context learning. Unlearning can be an effective control mechanism for the training phase, yet it does not prevent the model from performing an impermissible act during inference. We introduce a concept of \ununing, where unlearned knowledge gets reintroduced in-context, effectively rendering the model capable of behaving as if it knows the forgotten knowledge. As a result, we argue that content filtering for impermissible knowledge will be required and even exact unlearning schemes are not enough for effective content regulation. We discuss feasibility of \ununing for modern LLMs and examine broader implications.
\end{abstract}

\begin{document}

\maketitle

\section{Introduction}

Recent advancements in Large Language Models (LLMs) raise concerns about their use for undesirable purposes. Unlearning emerged as a promising solution for knowledge control, originally developed for removal of privacy-sensitive information~\citep{bourtoule2021sisa}. Since then, several works have attempted to utilize unlearning for a host of applications relating to the removal of undesired knowledge or behaviours: removing harmful capabilities~\citep{lynch2024eight} or harmful responses~\citep{yao2023large,liu2024towards}, erasing backdoors~\citep{liu2022backdoor} or specific information or knowledge pertaining to a particular topic ~\citep{eldan2023whos,li2024wmdp}, erasing copyrighted content~\citep{yao2023large} and even reducing hallucinations~\citep{yao2023large}. Such applications have been studied in the context of diffusion models too, with various attempts to use unlearning to remove unsafe concepts~\citep{zhang2023forget,fan2023salun}.

\noindent This paper discusses the application of unlearning to LLMs for removal of broadly impermissible knowledge, the use-case often discussed in policy circles e.g. for removal of biological and nuclear knowledge~\citep{li2024wmdp}. In fact, we uncover a fundamental inconsistency of the unlearning paradigm for this application. While unlearning aims to erase knowledge, the inherent \textit{in-context learning (ICL)} \citep{brown2020fewshotlearners,kossen2024incontext,agarwal2024manyshot} capabilities of LLMs introduce a major challenge. We introduce the concept of \ununing, where successfully unlearned knowledge can resurface through contextual interactions. This raises a critical question: if unlearned information can be readily reintroduced, is unlearning a truly effective approach for making sure that the model does not exhibit impermissible behaviours? We discuss the ramifications of~\ununing, particularly the need for effective content regulation mechanisms to prevent the resurgence of undesirable knowledge. Ultimately, we question the long-term viability of unlearning as a primary tool for content regulation.

\noindent {\color{red}Note}, this paper explicitly only considers the case when unlearning is used for purposes of content regulation i.e. problems formulated as \textit{as a model developer I do not want my model to be able to perform X}, where X can be e.g.~bioweapons development~\citep{li2024wmdp}. Entities deploying models operate under the expectation that those models don't pose a risk of being exploited for dangerous applications like weapons development. Importantly, it does not cover the original use-case of unlearning for the privacy purposes. 

\section{Nomenclature}

In what follows, we rely on six main terms: 

\noindent \textbf{(Informal) Definition 1. Knowledge} refers to information available to the model. This information can take up different forms and includes e.g. in-context provided inputs, information stored in the parameters of the model, or evidence available for retrieval. 

\noindent \textbf{(Informal) Definition 2. Content filtering} refers to the process of filtering out queries to and responses from a given model. Filtering can both be a part of the model, as well as, be external to it~\citep{glukhov2023llm}.

\noindent \textbf{(Informal) Definition 3. Unlearning} refers to a process in which knowledge is removed from a given model. This is a broad description that can encompasses different application scenarios. Below, we provide two informal definitions of unlearning.

\noindent \textbf{(Informal) Definition 4. Unlearning for privacy} seeks to remove knowledge that is defined as a particular subset of the model's original training dataset, referred to as the ``forget set''. Formal definitions ~\citep{ginart2019making,sekhari2021remember} require the (distribution of the) unlearned model to be indistinguishable from (the distribution of the) model retrained excluding the forget set. According to this notion, an unlearning method can either be exact~\citep{bourtoule2021sisa,muresanu2024unlearnable}, guaranteeing indistinguishability between the aforementioned distributions, or inexact~\citep{golatkar2020eternal,thudi2022algorithmic,kurmanji2024towards}, instead offering an approximation to that goal in exchange for greater efficiency or better model utility.

\noindent \textbf{(Informal) Definition 5. Unlearning for content regulation} seeks to remove knowledge that is (believed to be) associated with producing impermissible content. Note that the specification of this knowledge in this case does not necessarily take the form of identifying a subset of the training dataset. Instead, it captures more broadly information that we want to remove from the model (regardless of which subset of training data led to acquiring it), in order to prevent it from generating impermissible content. 
This notion aligns with the recent related notion of~\citet{goel2024corrective} for ``corrective unlearning''.

\noindent \textbf{(Informal) Definition 6. In-Context Learning} refers to an emergent capability of language models to generalise to tasks from task descriptions without these tasks present in the training data~\citep{brown2020fewshotlearners,milios-etal-2023-context,kossen2024incontext}. At the time of writing, such a model capability is ubiquitous, albeit some tasks are not perfectly solvable from descriptions alone.

\noindent For example, it may seem natural to remove any mention of what term \textit{bomb} means from the training set of the model, expecting that when asked to make bombs the model would fail -- simply because it possess no knowledge of what the word bomb means. Yet, if a term \textit{bomb}, for example under a different name as we show in Figure~\ref{fig:one_plot}, is defined as the a substance with certain properties and the model possess reasoning skills, a recipe for a bomb can be derived by the model, provided there is enough chemical knowledge still available to the model.

\noindent Importantly, we highlight that \textbf{even exact unlearning} would not stop the model from performing an impermissible act in the example above. That is because even though the model never saw the bomb-defining data, it still possesses all of the knowledge required to construct one. Consequently, no unlearning method can lead to `better removal' of the impermissible knowledge, compared to the model that never trained on this knowledge in the first place. We therefore argue that if \ununing poses an issue even in this `idealized unlearning' setting, this problem can only be exacerbated under imperfect unlearning. 

\noindent Given it is hard to reason about knowledge compositionality, i.e. how knowledge interacts with other knowledge available to  the model, it is not always obvious what logical inference given (benign) knowledge will enable. As such, attribution of model behaviour to specific knowledge that a user introduced is not always trivial. 

\section{Types of Knowledge}

\noindent To best understand the intuition behind~\ununing, we first discuss types of knowledge available to our models. We broadly classify knowledge into two main categories: \textbf{axioms} and \textbf{theorems} to represent facts and assumptions, and the derivative knowledge. 

\begin{figure*}
    \centering
    \includegraphics[width=0.8\linewidth]{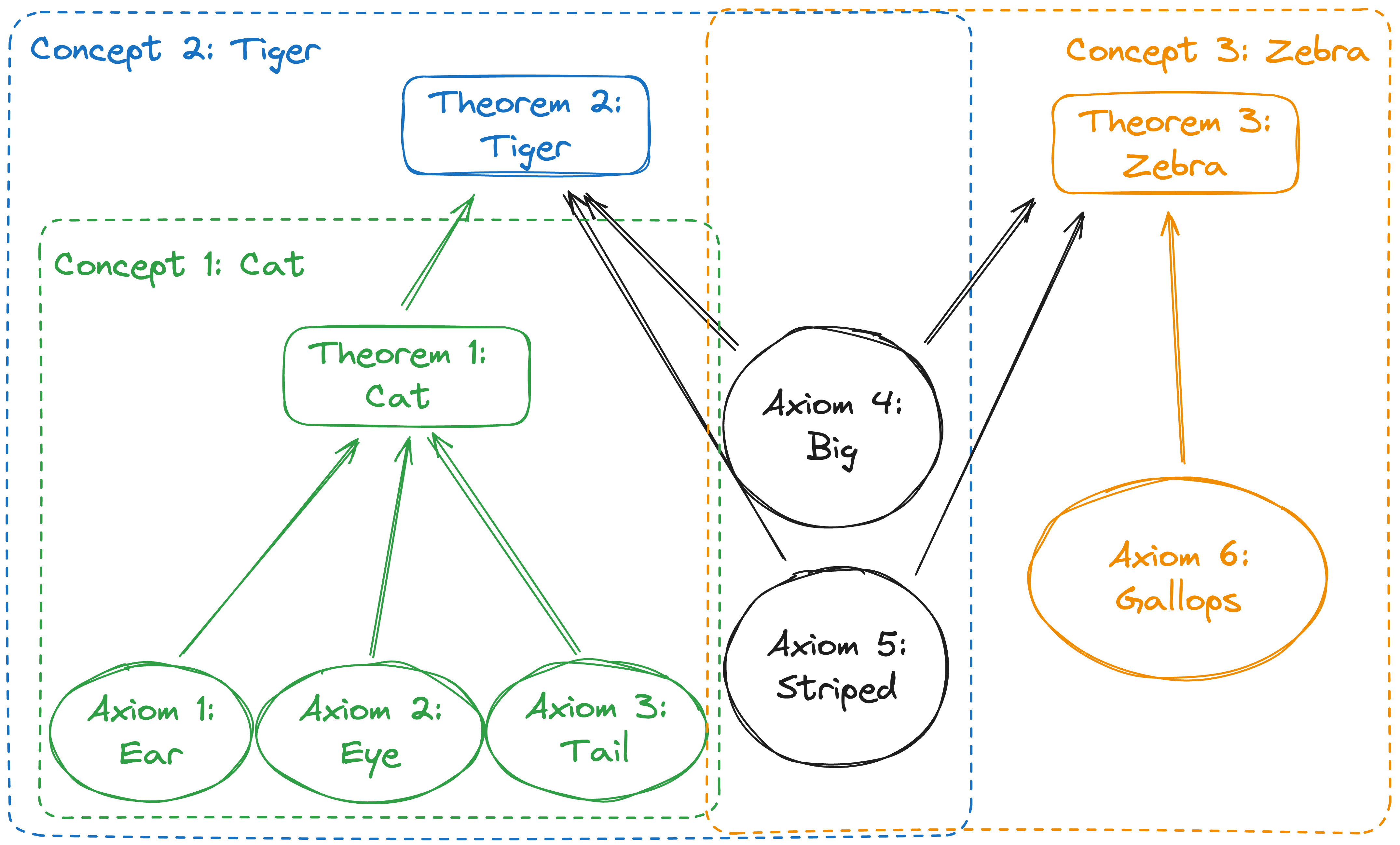}
    \caption{We broadly separate the knowledge into two main types: \textbf{axioms} and \textbf{theorems} to represent given facts and derived knowledge respectively. In the example above we assume that all theorems are defined in terms of underlying axioms, where some axioms are shared by different theorems. While unlearning of Cat may let the model forget what Cat means, it is relatively easy to redefine it provided that the underlying axioms are preserved. }
    \label{fig:knowledgetypes}
\end{figure*}

\noindent Consider the example depicted in Figure \ref{fig:knowledgetypes}, where we have three concepts defined out of 6 axioms. Here we define an axioms of Ear, Eye, Tail, Big, Striped, and Gallops as the smallest units of knowledge. We then use these to define three main theorems. If it has Ear, Eye, and Tail then we say its a Cat. If it is a Cat, but also is Big and Striped then it is a Tiger. Yet, if it is Big, Striped and Gallops then its a Zebra. 

\begin{figure*}
    \centering
    \includegraphics[width=0.9\linewidth]{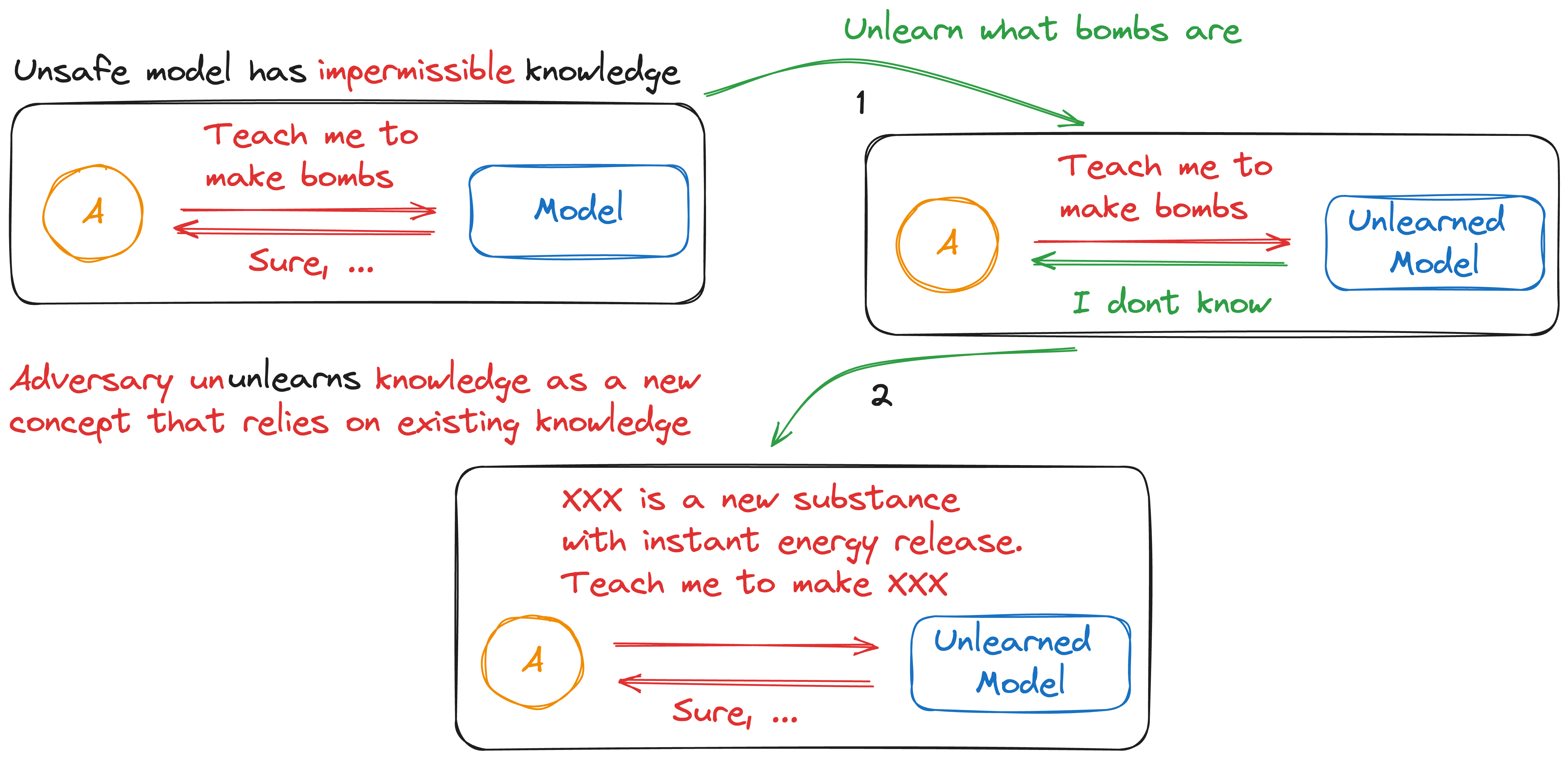}
    \caption{The figure demonstrates the concept of~\ununing. Here, the model that at first possess impermissible bomb making knowledge. The defender uses exact unlearning to remove all instances of usage of the term \textit{bomb}, making the model incapable of providing \textit{bomb} recipes, since it does not possess the knowledge of what that term describes. The adversary uses the knowledge still available to the model to describe the concept and as a result the model provides a response. }
    \label{fig:one_plot}
\end{figure*}

\noindent Now lets consider the case where concept of Tiger is impermissible and any queries regarding the Tiger are not allowed. In other words, the model developer explicitly expects that the model would not be used to reason about Tigers. We unlearn the concept of Tiger perfectly, using either exact unlearning or a very strong approximate unlearning scheme. Yet, even with the concept of Tiger gone, all of the underlying axiomatic knowledge is still retained inside of the model, since its features are used for other theorems available in the model, namely Zebra and Big in the example. If this knowledge were to also be unlearned from the model, it would start impacting the utility on other permissible tasks e.g. for reasoning about Zebra.

\noindent In the same way, many other conflicting objectives will leave the model possessing impermissible knowledge. For example, it may be desirable for the model to respond to  high-school chemistry essay questions, yet not to have  bomb-making knowledge. Figure~\ref{fig:one_plot} presents one such example. 

\section{\Unun}

\textbf{Setting:} Assume a model $M$ that takes in $x \in \mathcal{X}$ and outputs $\mathcal{Y}$. Note that here $\mathcal{X}$ includes both training and inference-time data. We assume we have an unlearning method $u$ that takes in a model and points $\hat{\mathcal{X}} \subseteq \mathcal{X}$ and outputs a model $\hat{M}$ that unlearned the said points $u(M, \hat{X}) = \hat{M}$. In this paper we assume that $\hat{\mathcal{X}}$ is the impermissible knowledge that is defined and identified by the model developer. We argue that by relying on in-context learning (ICL) an adversary can bring back the knowledge such that $\hat{M}$ when prompted with special context outputs the same results are $M$ over $\hat{\mathcal{X}}$. In other words, $M(\hat{X}) \approx \hat{M}(prompt+\hat{X})$.

\noindent \textbf{(Informal) Definition 7. \Unun} refers to a process where previously removed or never learned knowledge is instructed into the model using in-context learning. 

\noindent In this paper we argue that \ununing is a concern that should be taken into account when unlearning schemes are used and designed for removing impermissible functionality; as well as, models openly released. Note that \ununing applies even to exact unlearning. 

\section{Discussion}

There is a number of ramification of~\ununing.

\noindent \textbf{Need for Effective Filtering Mechanisms} \Unun~implies that for unlearning to remain effective, one needs to not only remove impermissible knowledge, but also perform a continuous, active process of suppressing in-context attempts to reintroduce that knowledge. Fundamental computational bounds suggest that such filtering is likely limited too, for example because of dependence on the context in which queries appear~\citep{glukhov2023llm}.

\noindent \textbf{Definitions and Mechanisms for Unlearning}
Given the above, is traditional unlearning by itself even a worthwhile pursuit for controlling the use of impermissible knowledge? Consider the Tiger--Zebra setting we discussed earlier. If we accept classic definitions for unlearning, a model that never trained on any Tiger knowledge represents perfect unlearning by construction. In other words, no unlearning method would perform better than a model that never saw any of the knowledge. Yet, since the model with in-context introduced concept of Tiger  can successfully reason about the Tiger it is clearly not confirming to the objective of not reasoning at all about it. Here, further exploration into the precise definitions and mechanisms for~\ununing is essential -- we need to find a way to explicitly limit the reasoning capability for our models that is invariant to prompting and other types of learning.

\noindent \textbf{Attributing knowledge} If a small benign axiom forces the attacker to discover a malicious theorem, who should be attributed for it? This dilemma mirrors the age-old philosophical debate – should an act be attributed to an individual who directly executes the act, the person who gives the order, the manufacturer of the tool, or the original tool designer~\citep{owen1992moral,Fischer1998responsibility}?
We argue that given that knowledge can be introduced by different parties, unlearning should not be assumed to be the sole mechanism for content policy enforcement.

\noindent \textbf{Forbidding knowledge} Instead of filtering out data, it may be better to teach the model explicitly that some knowledge is off limits~\citep{henderson2023self}. Do note however that this is not a bullet proof solution and will unlikely to be robust against, for example, mosaic attacks~\citep{glukhov2023llm}. Furthermore, the approach of forbidding knowledge requires foreseeing the types of harmful tasks or use-cases ahead of time and does not guarantee preventing harmful use more generally, beyond a given set of identified use-cases. Finally, it is not clear how a model should react to violations of its content regulations. In privacy literature it is a known fact that existence of the privacy mechanism can lead to an increased privacy leakage~\citep{wang2022differential}, and a similar effect was previously observed in unlearning~\citep{hayes2024inexact}. In other words, a model rejecting a request to synthesise a given chemical recipe can also help the malicious user understand what recipes can be used for malice. 

\section{Conclusion}

In this paper we argue that unlearning offers an incomplete solution for impermissible knowledge removal in LLMs with strong ICL capability. \Unun forces us to rethink unlearning as a one-size-fits-all solution and places emphasis on content filtering.

\bibliographystyle{abbrvnat}
\nobibliography*
\bibliography{template_refs}

\end{document}